\documentclass{article}

\PassOptionsToPackage{numbers, compress, sort}{natbib}

\usepackage[preprint]{neurips_2021}

\usepackage[utf8]{inputenc} %
\usepackage[T1]{fontenc}    %
\usepackage{booktabs}       %
\usepackage{amsfonts}       %
\usepackage{nicefrac}       %
\usepackage{microtype}      %
\usepackage{comment}
\usepackage{amssymb}
\usepackage{mathalpha}
\usepackage{float}
\usepackage{cuted}
\usepackage{multirow}

\usepackage[dvipsnames]{xcolor}
\usepackage{tabularx}
\usepackage{mathtools}
\usepackage{nonfloat}
\usepackage{enumitem}%
\usepackage{verbatim}
\usepackage{subcaption}
\usepackage{amsfonts} %
\DeclareMathSymbol{\shortminus}{\mathbin}{AMSa}{"39}
\DeclareFontFamily{OT1}{pzc}{}
\DeclareFontShape{OT1}{pzc}{m}{it}{<-> s * [1.150] pzcmi8t}{}
\DeclareMathAlphabet{\mathpzc}{OT1}{pzc}{m}{it}
\usepackage[mathscr]{eucal}
\usepackage{comment}
\usepackage{soul}
\usepackage{xfakebold}

\newcommand{\fbseries}{\unskip\setBold\aftergroup\unsetBold\aftergroup\ignorespaces}
\makeatletter
\newcommand{\setBoldness}[1]{\def\fake@bold{#1}}
\makeatother

\definecolor{cadmiumgreen}{rgb}{0.0, 0.42, 0.24}

\DeclareMathSymbol{\shortminus}{\mathbin}{AMSa}{"39}

\newcommand{\fig}[1]{Fig~\ref{fig:#1}}
\newcommand{\sect}[1]{Sect~\ref{sect:#1}}

\newcommand{\tab}[1]{Table~\ref{tab:#1}}

\newcommand{\eq}[1]{Eq. (\ref{eq:#1})}

\usepackage{xspace}

\usepackage[pagebackref=true,breaklinks=true,colorlinks,bookmarks=false]{hyperref}
\newcommand{\MethodName}{FLAME-\textit{in}-NeRF\xspace}

\newcommand{\xbf}{\mathbf{x}}

\newcommand{\gammabf}{\pmb{\gamma}}

\title{\includegraphics[height=15pt]{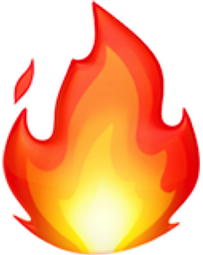} FLAME-\textit{in}-NeRF  \includegraphics[height=15pt]{Figures/fire.png}: Neural control of Radiance Fields for Free View Face Animation}

\author{%
  ShahRukh Athar \\
  Stony Brook University \\
  \texttt{sathar@cs.stonybrook.edu}
  \And
  Zhixin Shu \\
  Adobe Research \\
  \texttt{zshu@adobe.com} \\
   \And
   Dimitris Samaras \\
  Stony Brook University \\
  \texttt{samaras@cs.stonybrook.edu}
}

\begin{document}

\maketitle

\begin{abstract}

This paper presents a neural rendering method for controllable portrait video synthesis.
Recent advances in volumetric neural rendering, such as neural radiance fields (NeRF), has enabled the photorealistic novel view synthesis of static scenes with impressive results. However, modeling dynamic and controllable objects as part of a scene with such scene representations is still challenging. 
In this work, we design a system that enables both novel view synthesis for portrait video, including the human subject and the scene background, and explicit control of the facial expressions through a low-dimensional expression representation.
We leverage the expression space of a 3D morphable face model (3DMM) to represent the distribution of human facial expressions, and use it to condition the NeRF volumetric function.
Furthermore, we impose a spatial prior brought by 3DMM fitting to guide the network to learn disentangled control for  scene appearance and  facial actions.
We demonstrate the effectiveness of our method on free view synthesis of portrait videos with expression controls. To train a scene, our method only requires a short video of a subject captured by a mobile device.
\end{abstract}

\begin{figure*}[h]
    \centering
    \includegraphics[width=1.0\linewidth]{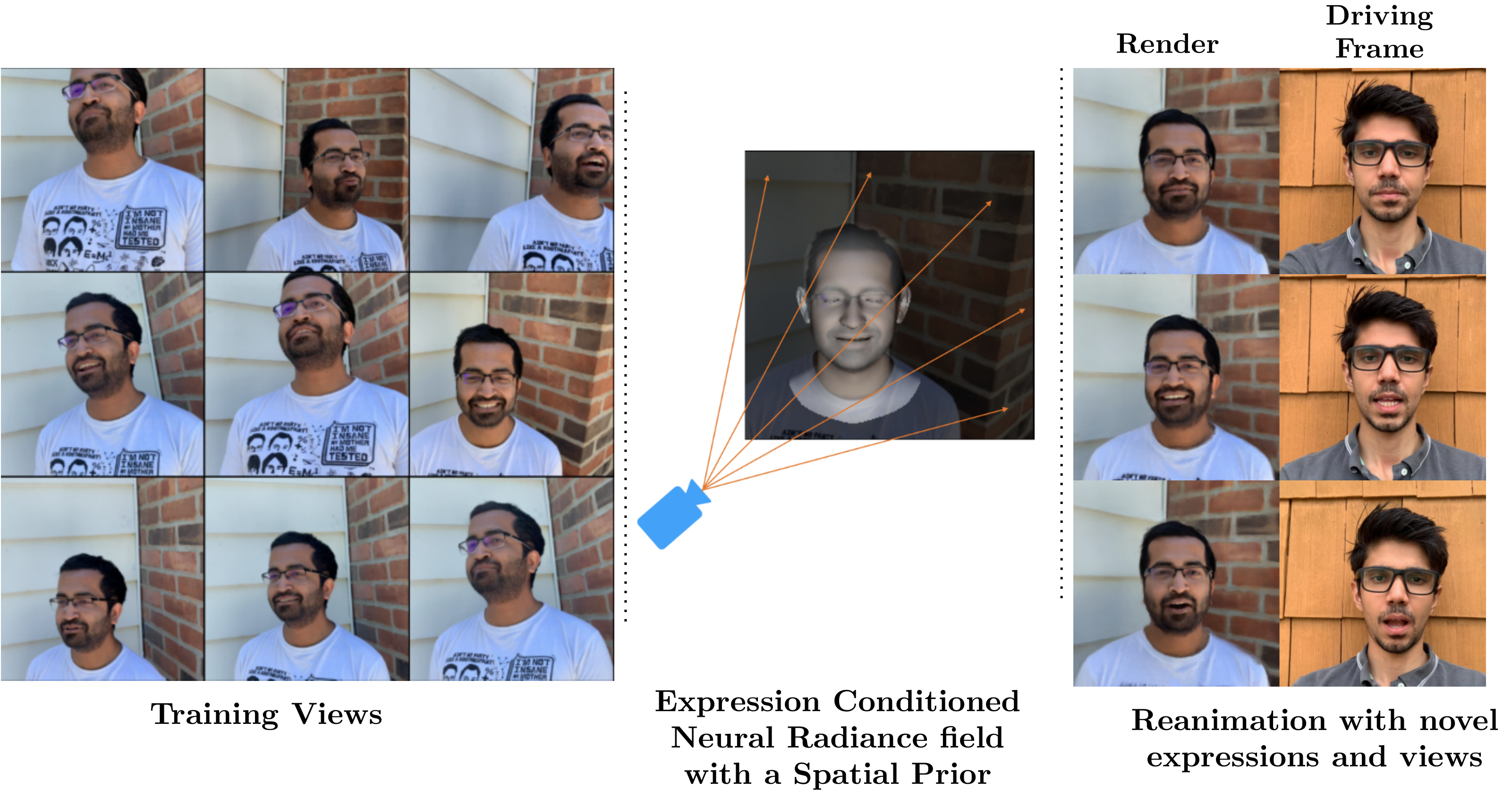}
    
    \caption{{\textbf{\MethodName.} Our method, \MethodName, models portrait videos (left) using an expression conditioned neural radiance field with a spatial prior (middle). Once trained, \MethodName can reanimate the subject and the scene present in the portrait video with arbitrary facial expressions and novel views. 
    }}
    \label{fig:teaser}
\end{figure*}

\section{Introduction}
A fully controllable human head model in natural scenes with arbitrary view synthesis still remains elusive, consequently, attracting immense interest in the Computer Vision, Machine Learning and Computer Graphics communities. Such a model, in principle, allows for arbitrary control of  human head pose, facial expression, identity and viewing direction. Earliest attempts towards a fully controllable human head model were in the form of 3D Morphable Models (3DMMs) \cite{blanz1999morphable}. 3DMMs use a PCA-based space to independently control face shape, facial expressions and appearance and can be rendered in any view using standard graphics-based rendering techniques such as rasterization or ray-tracing. However, 3DMMs \cite{blanz1999morphable} lack the ability to capture fine details of the human head such as hair, skin details and accessories such as glasses. Additionally, only the 3D face can be viewed in novel directions, the scene itself cannot, as the mesh only models the human head and nothing else. In contrast, recent methods for novel view synthesis of static and dynamic scenes \cite{nerf,nerfies,NerFACE, IDR, SRF, NeuralSceneFlow,DNeRF, SVS, kaizhang2020, unisurf, Gao-portraitnerf,sitzmann2019scene,Liu-2020-NSV, Zhang-2020-NAA,Bemana-2020-XIN,Martin-2020-NIT,xian2020space} are able to generate high quality novel views of a given captured scene but lack any control of the objects contained within the scene, including that of the human face and its various attributes.

In this paper we introduce \MethodName (pronounced Flamin-NeRF), a method that is capable of arbitrary facial expression control and novel view synthesis. We represent the whole scene as a neural radiance field in a manner similar to \cite{nerf,nerfies,NerFACE} and lend it explicit expression controls using expression parameters derived from a morphable model \cite{FLAME:SiggraphAsia2017}. Our model is trained on videos captured using a mobile phone, either by oneself or by someone else. In order to ensure only certain parts of the scene are influenced by the expression parameters, we, once again, utilize the 3DMM to impose a spatial prior on the 3D scene. Such prior ensures explicit disentanglement between appearance and expression in parts of the 3D scene where we know the human head is not present, ensuring that the appearance of scene points that do not project on the human face are unaffected by changes in expression. We show that not having such a disentanglement severely affects reanimation quality. Once trained, \MethodName allows for explicit control of both facial expression and viewing direction while capturing rich details of the scene \cite{nerfies, nerf} along with fine details of the human head such as the hair, beard, teeth and accessories such as glasses. Videos reanimated using our method maintain high fidelity to both the driving video in terms of facial expression manifestation and the original captured scene and human head. 

In summary, our contributions are as follows: 1) We propose a first-of-a-kind neural radiance field capable of explicit control on objects, such as the human face, within the captured scene. 2) We experimentally show the expression-appearance entanglement when reanimating portrait videos using neural radiance fields. We introduce a spatial ray sampling prior that ensures explicit disentanglement between facial expressions and appearance and significantly improves quality of reanimation. 3) We develop a system capable of simultaneous control of facial expressions and viewing direction trained on videos captured from a mobile phone.
\section{Related Work}
\MethodName is a method for arbitrary facial expression control and novel view synthesis of scenes captured in portrait videos. It is closely related to recent work on neural rendering and novel view synthesis, 3D face modeling, and controllable face generation. Below we discuss these related work.

\paragraph{Neural Scene Representations and Novel View Synthesis.} \MethodName is related to recent advances in neural rendering and novel view synthesis  \cite{nerf,nerfies,NerFACE, IDR, SRF, lassner2020pulsar, lombardi2021mixture, NeuralSceneFlow,DNeRF, SVS, kaizhang2020, unisurf, Gao-portraitnerf,sitzmann2019scene,Liu-2020-NSV, Zhang-2020-NAA,Bemana-2020-XIN,Martin-2020-NIT,xian2020space, Wizadwongsa2021NeX}. Notably, Neural Radiance Fields (NeRF) uses a Multi-Layer Perceptron (MLP), \(F\), to learn a volumetric representation of a scene. Given a 3D point and the direction from which the point is being viewed, \(F\) predicts its color and volume density. For any given camera pose, \(F\) is first evaluated densely enough throughout the scene using hierarchical volume sampling \cite{nerf}, then volume rendering is used to render the final image. \(F\) is trained by minimizing the error between the predicted color of a pixel and its ground truth value. While NeRF is able to generate high quality and photo-realistic images for novel view synthesis, it is only designed for a static scene and is unable to represent scene dynamics. Specifically designed for dynamic portrait video synthesis, our approach not only models the dynamics of human faces, but also allow specific control on the facial animation. 

\paragraph{Dynamic Neural Scene Representations.} Methods such as \cite{NeuralSceneFlow, li2021neural, DNeRF, xian2020space} extend NeRF to dynamic scenes by providing as input a time component and along with it imposing temporal constraints using scene flow \cite{NeuralSceneFlow, xian2020space} or by using a canonical frame \cite{DNeRF}. Similarly, Nerfies \cite{nerfies} too work with dynamic scenes by mapping to a canonical frame, however it assumes that the movement is small. \MethodName, like \cite{nerfies}, models the  portrait video by mapping to a canonical frame and assumes that the head motion in the video is small.

\paragraph{3D Face Modeling.} The landmark work by Blanz and Vetter in \cite{blanz1999morphable} on 3D Morphable Models (3DMMs) was among one of the first works to enable full control over the shape, expression and view of a 3D face. Given the camera pose and expression and shape parameters, a 3DMM can be rendered using standard rasterization techniques to give an image of the of the 3D face with the desired shape and the given expression. However, due to restrictions posed by a relatively limited representational power of the PCA space, 3DMMs often underfit real world human faces and are unable to model fine details. Further, the restriction of using a fixed mesh topology all but rules out the possibility of modelling any structures that are not modelled by it, such as the hair or accessories such as glasses. More recent methods such as Deferred Neural Rendering \cite{dnr}, use a coarse mesh, a high-dimensional neural texture map and a neural renderer, to generate photorealistic render of the human head. If the coarse mesh is controllable (like that of a 3DMM), the rendered image is too. Similarly,  Neural Point-Based Graphics \cite{NPBG}, uses high dimensional point cloud features along with a multi-resolution neural renderer to generate a photorealistic image. However, both these methods only work if the geometry, in the form of a coarse mesh \cite{dnr} or a point cloud \cite{NPBG} is given. Therefore, they cannot be used to model scenes as, often, we do not have its geometry. In contrast, \MethodName does not need any a-priori information about the scene geometry to synthesize novel views and yet retains control of the 3D face contained in the portrait video.

\paragraph{Controllable Face Generation.} The advent of adversarial training \cite{goodfellow2014generative}, cycle-consistency losses \cite{CycleGAN2017} and powerful convolutional architectures \cite{pix2pix2016} have made it possible to perform high quality facial expression editing just using 2D images \cite{shu2018deforming, NeuralFace2017, athar2020self, pumarola2020ganimation, StarGAN2018, starganv2}. However, since these methods are restricted to images and are often trained on frontal datasets, their quality degrades as the pose of the input image changes. Methods such as \cite{Kim-2018-DVP, Doukas2021Head2HeadDF, head2head2020, facedet3d} that use a 3DMM to reanimate faces. While being able to do so with great detail, they are unable to perform novel view synthesis as they do not model the geometry of the whole scene. In \cite{NerFACE}, the authors use neural radiance fields to provide full control on the head. However, they do not model the background and it is assumed to be static. In contrast, \MethodName provides full control over the facial expressions of the person captured in the portrait video and has to ability to synthesize novel views. However, \MethodName does not provide control over the head-pose and we leave that to future work.
\section{\MethodName}
\begin{figure}[h]
    \centering
    \includegraphics[width=1.0\linewidth]{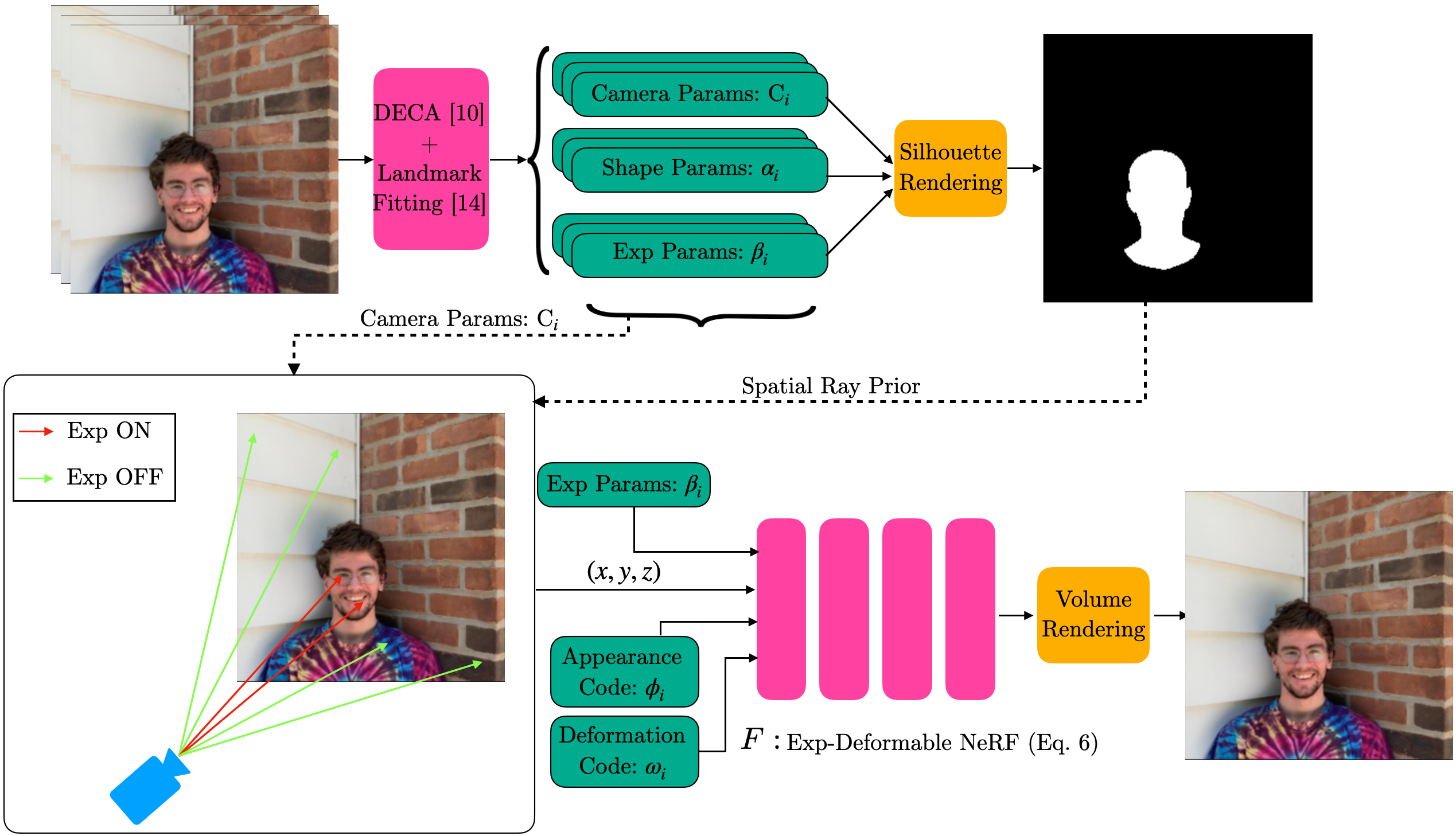}
    
    \caption{{\textbf{Overview of training \MethodName.} First, we use DECA \cite{DECA} and landmark fitting \cite{3DDFA_V2} to extract per-frame camera, shape, and expression parameters. Next, these parameters are used to render a silhouette of the FLAME model geometry. This silhouette is used to provide a spatial prior on ray sampling where only points that lie on rays that intersect the silhouette are affected by the expression parameters. Finally, given the \(i\)'th frame, we shoot rays, we sample points along them and input these points to the Deformable NeRF, \(F\), along with the \(i\)'th frame's expression parameters, deformation code and appearance code to render the final image.
    }}
    \label{fig:method}
\end{figure}

In this section, we describe our method, \MethodName, that enables novel view synthesis of dynamic portrait scene and arbitrary control of facial expressions. We model a dynamic portrait scene using a Neural Radiance Field with per-point deformation \cite{nerfies} to allow for small movement of the head. The deformation mechanism introduced in \cite{nerfies} deforms the rays of each frame to a canonical frame in order to ensure the rays that intersect are photometrically consistent. Facial expression dynamics are controlled by per-frame FLAME \cite{FLAME:SiggraphAsia2017} expression parameters derived using \cite{DECA} followed by standard landmarks fitting. In order to ensure disentanglement between the view parameters and the expression parameters, we adopt spatial prior on ray sampling during training. Specifically, we use a silhouette rendering of the fitted FLAME model \cite{FLAME:SiggraphAsia2017} and exclude the expression parameters for all points on rays that do not intersect the silhouette.

\subsection{Deformable Neural Radiance Fields}
A neural radiance field (NeRF) is a continuous function of, \(F: \left(\gammabf(\mathbf{x}), \gammabf(\mathbf{d})\right) \rightarrow (\mathbf{c}(\xbf, \mathbf{d}), \sigma(\xbf))\), that, given a 3D point of a scene \(\mathbf{x}\) and the viewing direction \(\mathbf{d}\) (i.e the direction of the ray it is on) gives the color \(\mathbf{c} = (r,g,b)\) and the density \(\sigma\). Here, \(F\) is a multi-layer perceptron (MLP) and \(\gammabf: \mathbb{R}^{3} \rightarrow \mathbb{R}^{3 + 6m}\) is the positional encoding \cite{nerf} defined as \(\gammabf(\xbf) = (\xbf,...,\text{sin}(2^{k}\xbf),\text{cos}(2^{k}\xbf),...)\) where \(m\) is the total number of frequency bands and \(k \in \{0,...,m-1\}\). The expected color of a camera ray \(\mathbf{r}(t) = \mathbf{o} + t\mathbf{d}\), where \(\mathbf{o}\) is the camera center and \(\mathbf{d}\) is the direction of the ray, is given by the standard volumetric rendering equation
\begin{equation}
    C(\mathbf{r})=\int_{t_{n}}^{t_{f}} T(t) \sigma(\mathbf{r}(t)) \mathbf{c}(\mathbf{r}(t), \mathbf{d}) d t, \text { where } T(t)=\exp \left(-\int_{t_{n}}^{t} \sigma(\mathbf{r}(s)) d s\right)
    \label{eq:vol_render}
\end{equation}
where, \(T(t)\) is the accumulated transmittance along the ray from \(t_{n}\) to \(t\). In practice, the integral in \eq{vol_render} is estimated using hierarchical volume sampling, we refer the reader to \cite{nerf} for details. Given multiple images of a scene, with their associated camera intrinsics and extrinsics, rays are shot through each pixel of each image of the scene using mini-batches. The color of each ray is accumulated via volume rendering using \eq{vol_render} and the error w.r.t the ground truth pixel color is minimized as follows:
\begin{equation}
    \underset{\theta}{\text{min}} \sum_{p}|| C_{p}(\theta) -  C_{p}^{GT} ||
    \label{eq:nerf_opt}
\end{equation}
where, \(p\) is an indexing variable over all the pixels in the mini-batch, \(\theta\) are the parameters of \(F\) and \(C_{p}^{GT}\) is the ground truth pixel color.

NeRF, as defined above, is naturally designed for static scenes where all the views of the scene are captured at the same time as it assumes that two rays that intersect would have the same color. However, as observed in \cite{nerfies} humans rarely remain perfectly static during a capture process, more so when they're speaking or performing facial expressions, thus vanilla NeRF training fails to model such dynamic scenes. In order to take into account for subtle movement of subjects in the capturing process, \cite{nerfies} proposed the Deformable NeRF architecture. In \cite{nerfies}, 3D points, \(\xbf\)'s, captured in the \(i\)'th frame of the video are deformed to a canonical space via a deformation function \(D_{i}: \xbf \rightarrow \xbf'\). Here,  \(D_{i}\) is defined as \(D(\xbf, \omega_{i}) = \xbf'\) where \(\omega_{i}\) is a per-frame latent deformation code. In practice, \(D(\xbf, \omega_{i})\) is modeled using an MLP and its coordinate input is also positionally encoded, we choose to omit it for brevity. In addition to a deformation code, \(\omega_{i}\), \cite{nerfies} also uses a per-frame appearance code, \(\phi_{i}\), thus the final radiance field for the \(i\)'th frame is as follows:
\begin{equation}
    F: \left(\gammabf(D(\mathbf{x},\omega_{i})), \gammabf(\mathbf{d}), \phi_{i}\right) \rightarrow (\mathbf{c}(\xbf, \mathbf{d}), \sigma(\xbf))
    \label{eq:nerfie_def}
\end{equation}
Now, in addition to the parameters of \(F\) each \(\omega_{i}\) and \(\phi_{i}\) are also optimized through stochastic gradient descent. In practice, \(D(\xbf, \omega_{i})\) is modelled as a dense SE(3) field, we refer the reader to \cite{nerfies} for details. While the aforementioned methods are able generate novel views \cite{nerf, nerfies} and handle small movement of objects in the scene \cite{nerfies}, they are still unable to control them.

\paragraph{Coarse-to-fine deformation regularization}
Inspired by \cite{nerfies}, we use coarse-to-fine regularization on the coordinate input of the deformation network, \(D\). Coarse-to-fine regularization is implemented by linearly increasing the weight of larger frequencies of the positional encoding starting zero. More specifically, the weight of frequency band \(l\) is:
\begin{equation}
    w_{l}(\alpha)=\frac{(1-\cos (\pi \operatorname{clamp}(\alpha(t)-l, 0,1))}{2}; \text{ where } \alpha(t) = \frac{mt}{N}
\end{equation}
where, \(m\) is the number of frquency bands in the positional encoding, \(t\) is the iteration number and \(N\) is a user-defined hyperparameter for the iteration after which all the frequency bands must be used. 
Coarse-to-fine regularization ensures that in the initial stages of training, the deformations do not become too large such that they can hurt generalization ability to novel views.

\subsection{Expression Control in Deformable Neural Radiance Fields}
\MethodName models changes of subject's facial expression as changes in the color of 3D scene points. To this end, we condition the learning of deformable NeRF on a set of expression parameters provided by FLAME \cite{FLAME:SiggraphAsia2017}. The expression conditioned deformable NeRF is defined as follows:
\begin{equation}
    F: \left(\gammabf(D(\mathbf{x},\omega_{i})), \gammabf(\mathbf{d}), \phi_{i}, \beta_{i}\right) \rightarrow (\mathbf{c}(\xbf, \mathbf{d}), \sigma(\xbf))
    \label{eq:flamin_nerf_no_cond}
\end{equation}
where, \(\beta_{i}\) is the expression parameter of the \(i\)'th frame, \(D\) is the deformation function, \(\omega_{i}\) and \(\phi_{i}\) are the deformation code and appearance code of the \(i\)'th frame respectively. The expected colors of each pixel are then calculated using \eq{vol_render}.

\subsection{Spatial Prior for Ray Sampling}
\label{sect:flame_spatial_prior}
\begin{figure}[h]
    \centering
    \includegraphics[width=1.\linewidth]{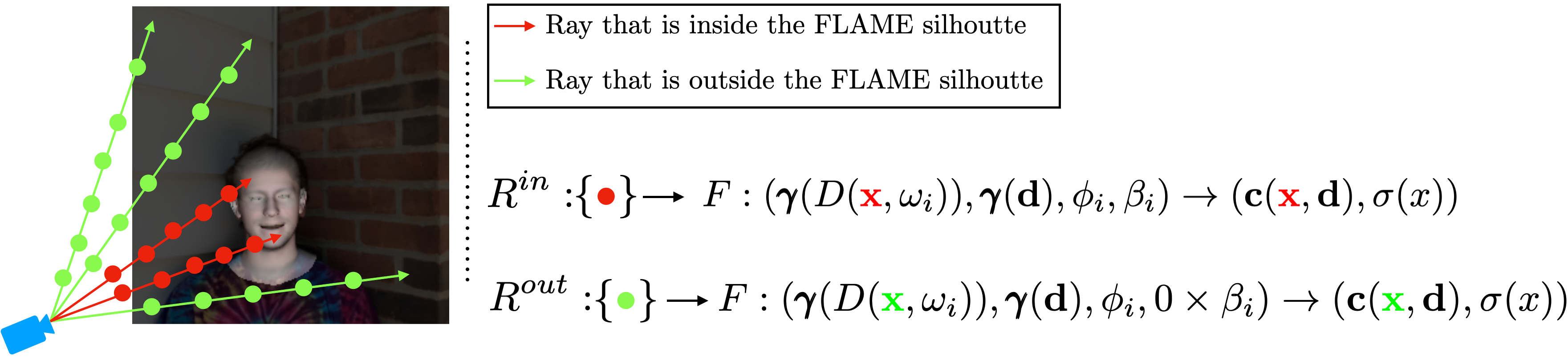}
    \caption{{\textbf{FLAME induced  Prior.} \MethodName uses a silhouette rendering of the FLAME model geometry (an overlay is shown above) to provide a spatial prior on rays shot through the 3D scene. All points that lie on rays that are intersect the silhouette,  shown in red, are affected by the expression parameters. Other points, shown in green, have their expression parameters set to zero and are therefore unaffected by changes in expression. 
    }}
    \label{fig:masking}
\end{figure}
\begin{figure}[h]
    \centering
    \includegraphics[width=1.0\linewidth]{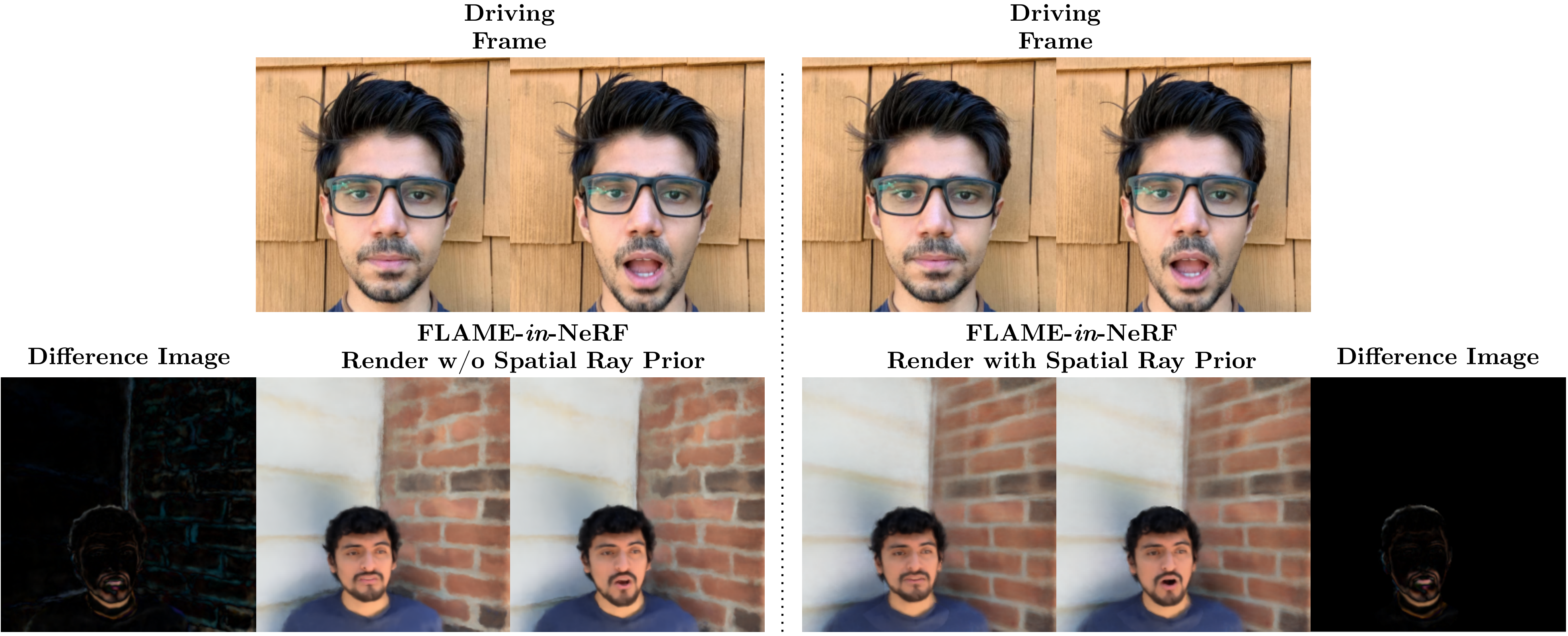}
    
    \caption{{\textbf{Why use the Spatial Ray Prior?}: Here we demonstrate the necessity of using a Spatial Ray Prior for the reanimation of portrait videos with arbitrary facial expressions and view control. On the left we have a model that does not use a Spatial Ray Prior and on the right, a model that does. As can be seen, the model without the prior generates results of lower quality (e.g. on the lines of the brick wall) than the model with it. Further, the difference images show that, \textit{despite keeping the viewing direction constant}, the model without the spatial prior changes the background appearance with changing expression. In contrast, the model with the spatial ray prior does not do so as the prior explicitly disentangles the expression and the appearance in regions of the 3D scene that do not project to the face. \textit{(Please watch the accompanying video in Supplementary)}. 
    }}
    \label{fig:Ablation}
\end{figure}

When the radiance field is modeled as described in \eq{flamin_nerf_no_cond}, there is nothing that prevents the appearance of a point \(\xbf\), that does \textit{not} project on the face, to become dependent on the expression parameters \(\beta_{i}\). In \sect{spatial_prior}, we show this phenomena of background expression dependence in practice; a radiance field that changes the appearance of points on the background as the view is kept constant but the expression changes (see \fig{Ablation}). In order to counter this effect, we use a spatial prior on rays that's induced by the fitted FLAME 3DMM. First, we render the FLAME mesh giving us a binary silhouette image \(S_{i}\) for each frame \(i\). Next, we define two sets of points, let \(R_{i}^{\text{in}} = {\xbf_{1}, ..., \xbf_{n}}\) be the set of points that lie on rays inside the FLAME silhouette of frame \(i\) and \(R_{i}^{\text{out}} = {\xbf_{1}, ..., \xbf_{m}}\) be the set of points that lie on rays outside the FLAME silhouette of frame \(i\) (as shown in \fig{masking}), the radiance field is now defined as: 
\begin{equation}
    F: \left(\gammabf(D(\mathbf{x},\omega_{i})), \gammabf(\mathbf{d}), \phi_{i}, \mathbb{I}(\xbf)\beta_{i}\right) \rightarrow (\mathbf{c}(\xbf, \mathbf{d}), \sigma(\xbf))
    \label{eq:flamin_nerf_cond}
\end{equation}
where, \(\mathbb{I}(\xbf) = 1 \text{ if } \xbf \in R_{i}^{\text{in}} \text{ and } 0 \text{ otherwise }\). This ensures that points that do not affect face pixels are not affected by facial expression changes. As can be seen in \fig{Ablation}, such a spatial ray prior effectively disentangles the appearance and expression and ensures that the background is unaffected by facial expression parameters.

\paragraph{Face region regularization.} Since our method optimizes extrinsic camera parameters w.r.t the FLAME 3DMM, we assume the head is static and has an identity mapping to the canonical frame. In order to prevent the deformation network, \(D\), from moving the 3DMM, we penalize any deformation on points sampled from it as follows:
\begin{equation}
   \underset{\psi,\omega_{i}}{\text{min}} ||D(\xbf_{\text{3DMM}}; \omega_{i}) - \xbf_{\text{3DMM}}||; \quad \forall i
\end{equation}
where, \(\psi\) are the parameters of \(D\).

\section{Results}

In this section, we show results of facial expression control and novel view synthesis on portrait videos captured using a standard smartphone.  First, we conduct an ablation showing the necessity of a spatial prior on ray sampling to train controllable neural radiance fields. Next, we compare \MethodName to Nerfies \cite{nerfies} which is the current state-of-the-art in novel view synthesis of portrait videos. We show a quantitative comparison with Nerfies \cite{nerfies} on validation data and a qualitative comparison as we drive learned neural radiance field using expression parameters extracted from a driving video. We use the deformation and appearance code of the first frame, for both methods, to perform the reanimation. Full videos of the reanimation can be found in the supplementary material. We strongly urge the readers to check them out to see \MethodName performing at its best.

\subsection{Training Data Capture and Training details}
The training data was captured using an iPhone XR smartphone for all the experiments in the paper. We ask the subject to enact a wide range of expressions and speech while trying to keep their head still as the camera is panned around them.  Alternatively, the subject can self-capture a video (a selfie video) as they speak. We calculate the expression and shape parameters of each frames in the videos using DECA \cite{DECA}. Next, we compute the extrinsic camera parameters via standard landmark fitting using landmarks predicted by \cite{3DDFA_V2}. All training videos are about just 25 seconds long (\(\sim\) 700 frames)%
. Due to compute restrictions, the video is down-sampled and the models are trained at 256x256 resolution. We use coarse-to-fine regularization \cite{nerfies} to train the deformation network \(D(\xbf, \omega_{i})\). Please find full details of each experiment in the \sect{exp_config}.

\begin{figure}[h]
    \hspace*{-1cm}
    \includegraphics[width=1.05\linewidth]{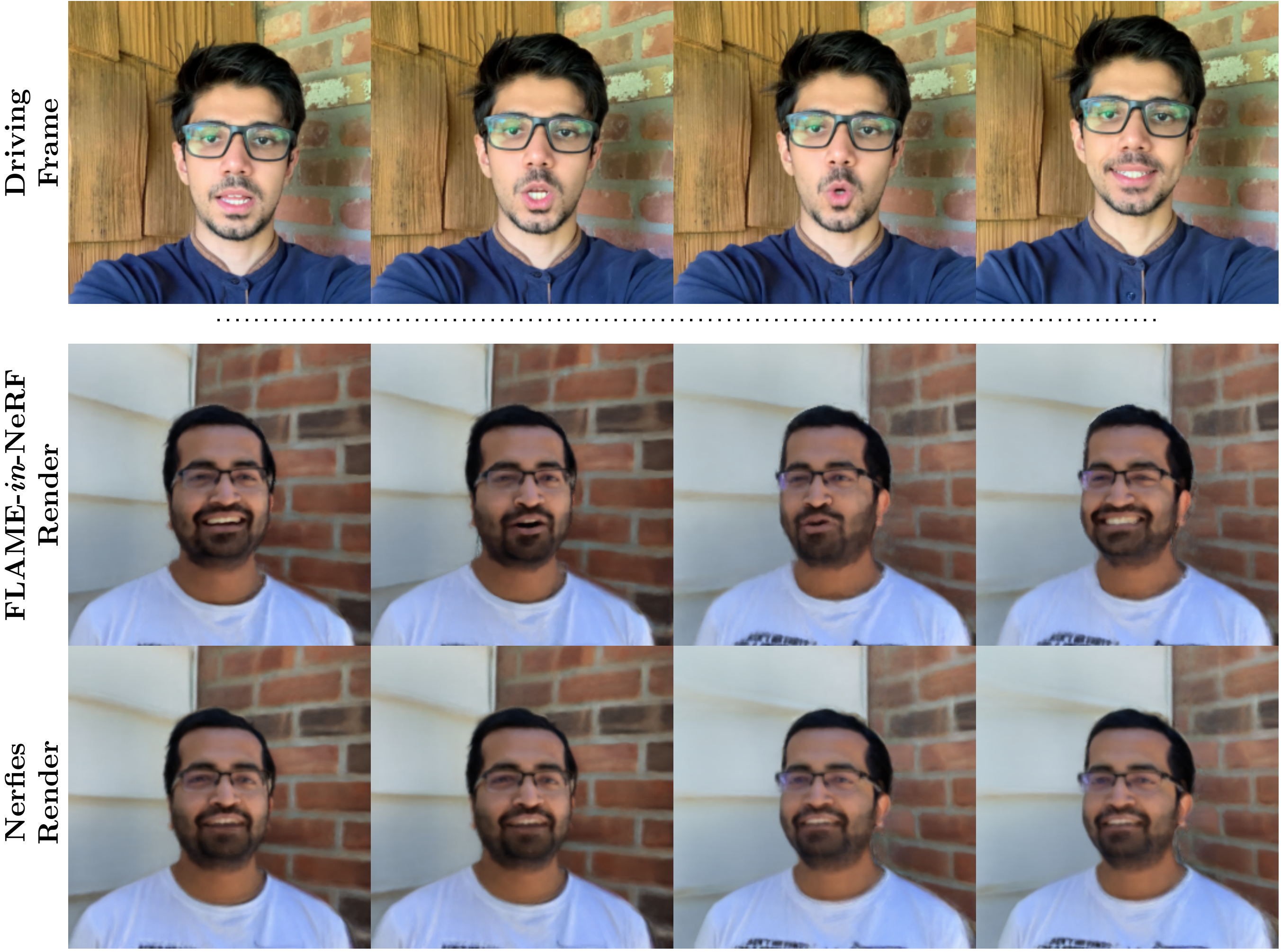}
    
    \caption{{\textbf{Reanimating Subject 1 using \MethodName}: Here we show the results of reanimating Subject 1 using both \MethodName and Nerfies \cite{nerfies} with both expression and view changes. The first row shows the driving frame, the second row shows the results of \MethodName and the third shows the results of Nerfies \cite{nerfies}. We see that \MethodName generates high-quality reanimation results with high fidelity to the driving expression and consistency across views. In contrast, Nerfies \cite{nerfies} is unable to model expression changes and produces lower quality results.
    }}
    \label{fig:Driving_Subj_1}
\end{figure}

\begin{figure}[h]
    \hspace*{-1cm}
    \includegraphics[width=1.05\linewidth]{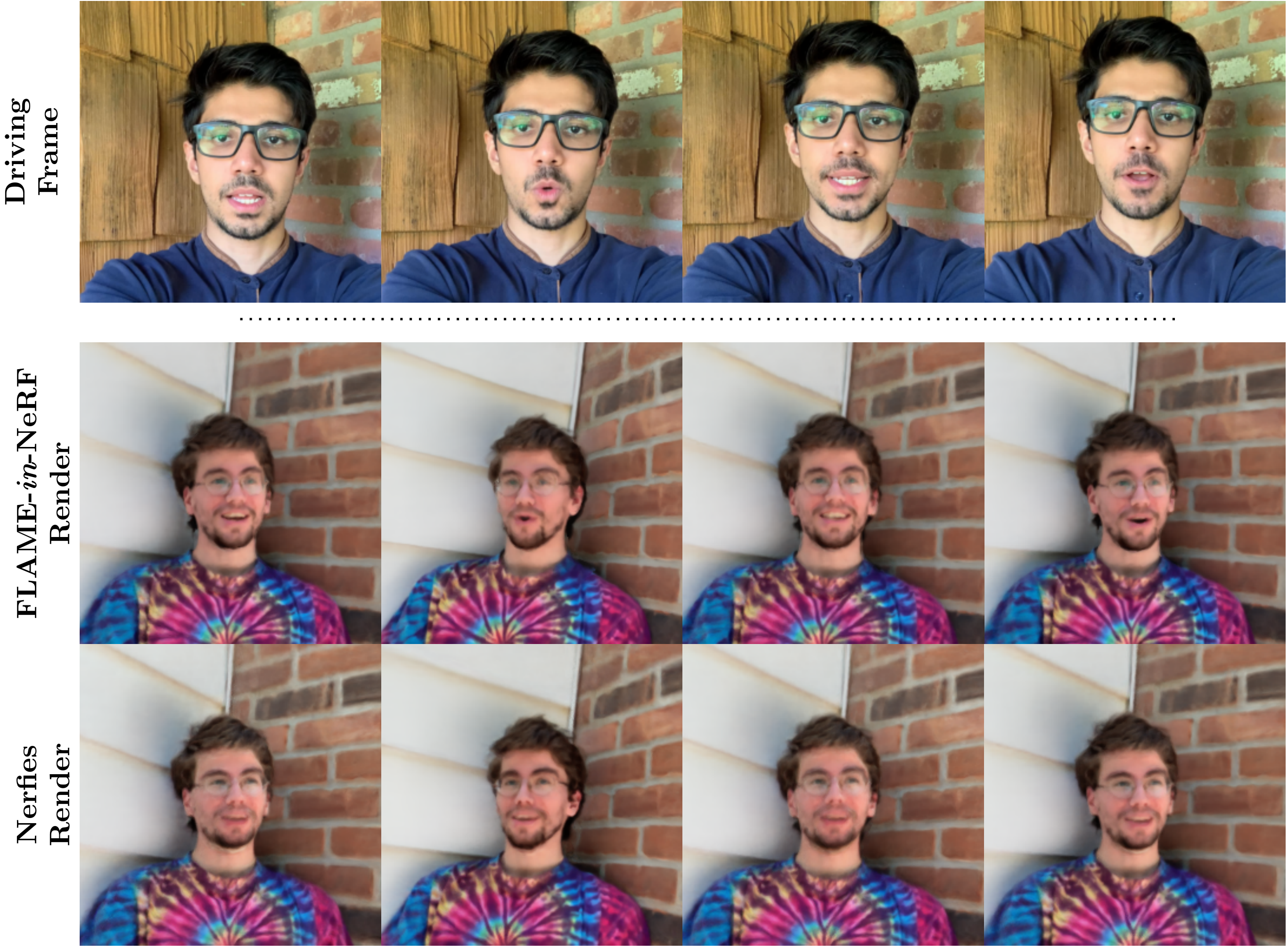}
    
    \caption{{\textbf{Reanimating Subject 2 using \MethodName}: Here we show the results of reanimating Subject 2 using both \MethodName and Nerfies \cite{nerfies} with both expression and view changes. The first row shows the driving frame, the second row shows the results of \MethodName and the third shows the results of Nerfies \cite{nerfies}. We see that \MethodName generates high-quality reanimation results with high fidelity to the driving expression and consistency across views. Nerfies \cite{nerfies}, while generating results that are quite consistent across views is unable to faithfully reproduce the expression of the driving frame.
    }}
    \label{fig:Driving_Subj_2}
\end{figure}

\subsection{On the necessity of a Spatial Ray Prior}
\label{sect:spatial_prior}
In this section we demonstrate the necessity of the FLAME induced spatial prior on ray sampling as discussed in \sect{flame_spatial_prior}. In \fig{Ablation}, we show the results of renanimating a portrait video with a constant view directions using a \MethodName with and without a spatial ray prior. As can be seen in \fig{Ablation}, the results of the model without the prior are of lower quality than that of the model with it. Additionally, when we calculate the difference image of the reanimated frames generated by both methods we see that, \textit{despite the viewing direction remaining constant}, the model without the prior changes the background. In stark contrast, and unsurprisingly, the model with the spatial ray prior does not do so and only makes changes around the face. The prior explicitly disentangles the expression parameters and the appearance in regions of the scene which do not project to the face. Therefore, we see that the spatial prior is a necessary ingredient high quality portrait video reanimation.

\begin{table}[]
\begin{tabular}{@{}llll@{}}
\toprule
Subject                     &Method                     & MSE (\(\downarrow\))     & PSNR (\(\uparrow\))  \\ \midrule
\multirow{2}{*}{Subject 1} &\MethodName & \hl{2.045e-3} & \hl{26.89} \\
&Nerfies                    & 2.82e-3  & 25.49 \\ \bottomrule
\multirow{2}{*}{Subject 2} &\MethodName & \hl{1.255e-3} & \hl{29.01} \\
&Nerfies                    & 2.05e-3  & 26.06 \\ \bottomrule
\end{tabular}
\caption{Quantitative results of Subject 1 and Subject 2 on validation data. Our results are significantly better than Nerfies \cite{nerfies} across both subjects.}
\label{tab:Subjects_metrics}
\end{table}

\subsection{Evaluation on Validation Data}
\label{sect:eval_data}
We evaluate both \MethodName and Nerfies \cite{nerfies} on held out images. Since both these methods use a per-frame deformation and appearance code, \(\omega_{i}\) and \(\phi_{i}\) respectively, we cannot perform a direct comparison with the ground truth image as it may have a different deformation to the canonical frame than the first frame (which is what we use as default for reanimation). Therefore, we first find the deformation of a given validation image to the canonical frame by optimizing the deformation code as follows
\begin{equation}
    \underset{\omega}{\text{min}} || C_{p}(\omega; {\xbf}, \mathbf{d}, \theta, \phi_{0}) -  C_{p}^{GT} ||
    \label{eq:def_opt}
\end{equation}
where, \(C_{p}(\omega; {\xbf}, \mathbf{d}, \theta, \phi_{0})\) is the predicted color at pixel \(p\) generated using \eq{vol_render} and \eq{flamin_nerf_cond}, \(\phi_{0}\) is the appearance code of the first frame, \(\theta\) are the parameters of \(F\) as defined in \eq{flamin_nerf_cond} and \(C_{p}^{GT}\) is the ground-truth pixel value. Note, we \textit{only} optimize \(\omega\), all other parameters of the radiance field are kept fixed. We optimize \eq{def_opt} for 2000 epochs which we observe to be more than enough to find the loss plateau. Once the optimization finishes, we report the final MSE and PSNR. As can be seen in \tab{Subjects_metrics}, our method out performs Nerfies \cite{nerfies} on validation images. Since we model with dynamic portrait videos with changing facial expressions, Nerfies is unable to learn the topological changes of the mouth, often regressing to a `mean' expression (see \fig{Driving_Subj_1} and \fig{Driving_Subj_2}) with small view-dependent changes. In contrast, with the expression conditioning the use of a spatial prior, \MethodName is able to model facial expressions with high fidelity thus giving better reconstructions.

\subsection{Reanimation with Arbitrary Expression Control and Novel View Synthesis}
\label{sect:reanim}
In this section we show results of reanimating Neural Radiance Fields using both \MethodName and Nerfies \cite{nerfies} using expression parameters as the driving parameters. Per-frame expression parameters from the driving video are extracted using DECA \cite{DECA} and are given as input to \MethodName as follows:

\begin{equation}
    F: \left(\gammabf(D(\mathbf{x},\omega_{i})), \gammabf(\mathbf{d}), \phi_{i}, \mathbb{I}(\xbf)\beta_{i}^{\text{Drive}}\right) \rightarrow (\mathbf{c}(\xbf, \mathbf{d}), \sigma(\xbf))
    \label{eq:flamin_nerf_cond_drive}
\end{equation}
 where, \(\beta_{i}^{\text{Drive}}\) are the expression parameters derived from the driving video. We refer the reader to \eq{flamin_nerf_cond} for the definitions of the other variables. Since Nerfies \cite{nerfies} do not take as input expression parameters, it's forward pass remains the same as in \eq{nerfie_def}. As we drive both methods, we also simultaneously change the viewing direction. \fig{Driving_Subj_1} shows the results of both \MethodName and Nerfies \cite{nerfies} with changing the expression parameters and view on subject 1. As one can see, \MethodName captures the driving expression with high fidelity and is able to do so regardless of viewing direction. Nerfies \cite{nerfies}, lacking the explicit conditioning on expression parameters, is unable to generate images with the correct facial expression. Only small changes in expression can be observed due to view changes. Similarly, \fig{Driving_Subj_2} shows the results of changing expression parameters and view on subject 2. As mentioned earlier, \MethodName is able to generate expressions with high fidelity regardless of view while Nerfies \cite{nerfies} is only able to capture changes in expression as view dependent effects.
\section{Conclusion}

In this paper we have presented \MethodName, a novel method capable of arbitrary facial expression control and novel view synthesis for portrait videos. \MethodName uses an expression-conditioned neural radiance field along with a spatial prior to generate images with high fidelity to both the subject in the original portrait video and the provided expression parameters, in any viewing direction. \MethodName is also able to model details of the subject's face such as hair and glasses and reproduce them with high fidelity as the video is driven. Additionally, \MethodName does not require any complex equipment to capture the portrait video, any commodity smartphone with a camera will do. However, the problem of controllable human head models with novel view synthesis is far from solved. \MethodName is unable to model large head movements and requires the subject in the portrait video to remain relatively still. We hope to address this in future work. These are exciting times to be a part of ML/CV/CG communities as neural methods push the state of the art in head control and novel view synthesis with broader impacts in entertainment, education and HCI.  However, our paper generates realistic manipulated videos, that can be used for fraudulent or misinformation purposes which would be a potential negative social impact shared with most face editing methods. Any methods that try to mitigate potential biases in 3D Morphable Models would also address bias concerns for our paper.

{\small
\bibliographystyle{abbrv}
\bibliography{refs}
}

\newpage
\appendix
\section{Supplementary}

\begin{figure*}[h!]
    \centering
    \includegraphics[width=1.0\linewidth]{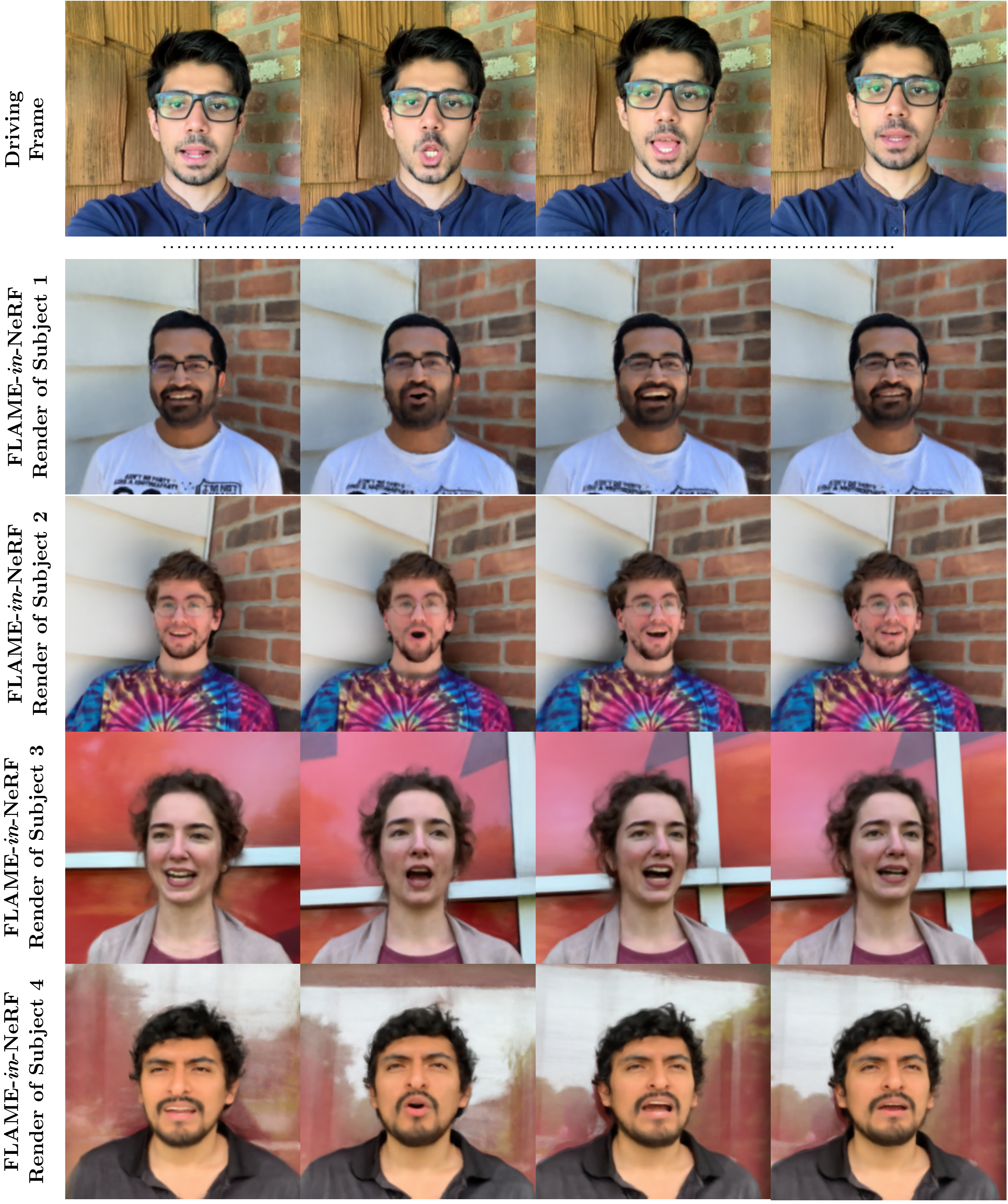}
    
    \caption{{\textbf{Reanimation using \MethodName.} Results of reanimating 4 subjects using a self-captured video (\href{https://drive.google.com/file/d/1-bmWJ8gBKyPUnUViUAVtFeWy7F22lgO1/view?usp=sharing}{Subject 1}, \href{https://drive.google.com/file/d/1AiSy2M7emLMtfWeSnWBenW2ItfCtVhv4/view?usp=sharing}{Subject 2}, \href{https://drive.google.com/file/d/1_Rorr1uvL9paD6JZ2_2T0yFLcNSU3b5i/view?usp=sharing}{Subject 3},\href{https://drive.google.com/file/d/10aIndNUQ79TNsejosviDE5qDHw7aRBdQ/view?usp=sharing}{Subject 4} ). The reanimated frames retain high fidelity to the target expression of the driving frame while, while simultaneously, respecting the individual characteristics of each subject. For example, in column 2, the rounding of the mouth is faithfully rendered across all the subjects but is individualistic. Subject 3 (in column 2) has her teeth showing as her mouth is rounded, while the others do not. Similarly, the half-open mouth of the last column is also faithfully rendered across all subjects while retaining individual characteristics.
    }}
    \label{fig:reanim_1}
\end{figure*}

\subsection{More Reanimation Results}
\begin{figure*}[h!]
    \centering
    \includegraphics[width=1.0\linewidth]{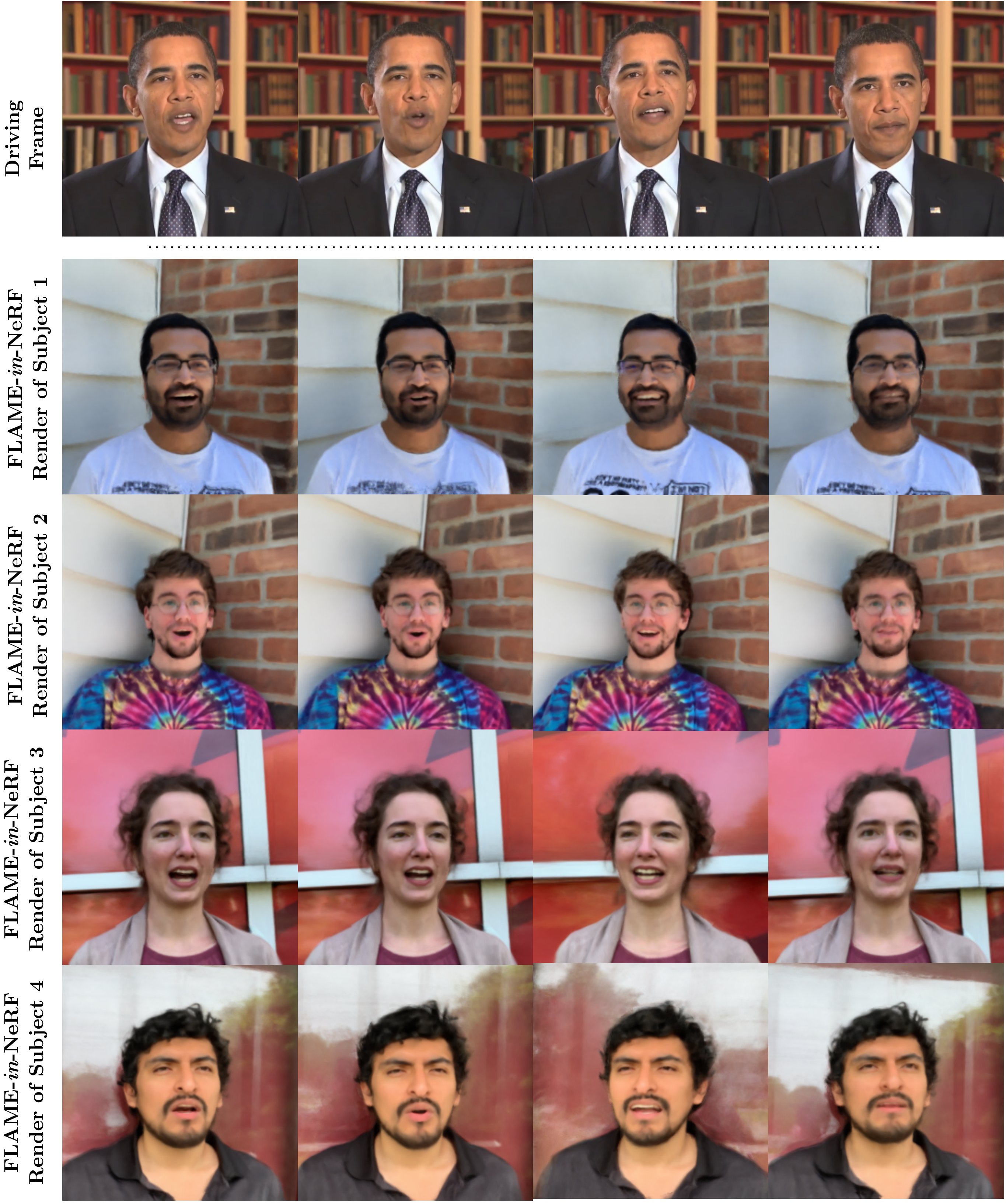}
    
    \caption{{\textbf{Reanimation using \MethodName.} Results of reanimating 4 subjects using a video from the internet. Despite being an in-the-wild video with a wide variety of expressions, the reanimated frames retain high fidelity to the target expression of the driving frame while, while simultaneously, respecting the individual characteristics of each subject. For example, in column 1, the half open mouth is faithfully rendered across all the subjects but is individualistic. Subjects 1 and 3 (in column 1) have their teeth showing prominently while Subjects 2 and 4 do not. 
    }}
    \label{fig:reanim_2}
\end{figure*}

\begin{figure*}[h!]
    \centering
    \includegraphics[width=1.0\linewidth]{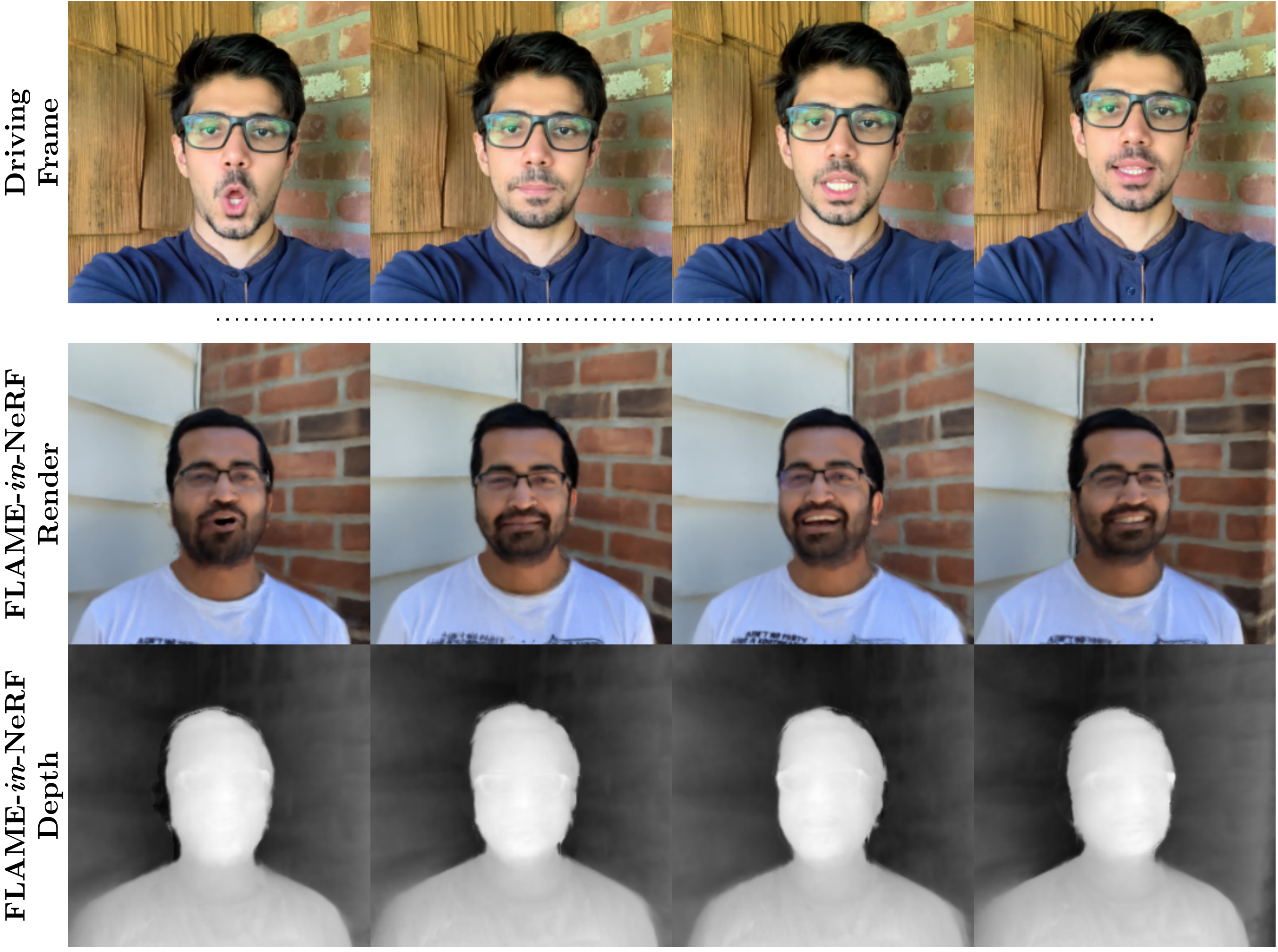}
    
    \caption{{\textbf{Reanimation with depth using \MethodName.} Results of reanimating Subject 1 using a self captured video. In the second row we show the reanimated frames and in the last row we show the rendered depth.   
    }}
    \label{fig:reanim_3}
    \vspace{-5mm}
\end{figure*}

\begin{figure*}[h!]
    \centering
    \includegraphics[width=1.0\linewidth]{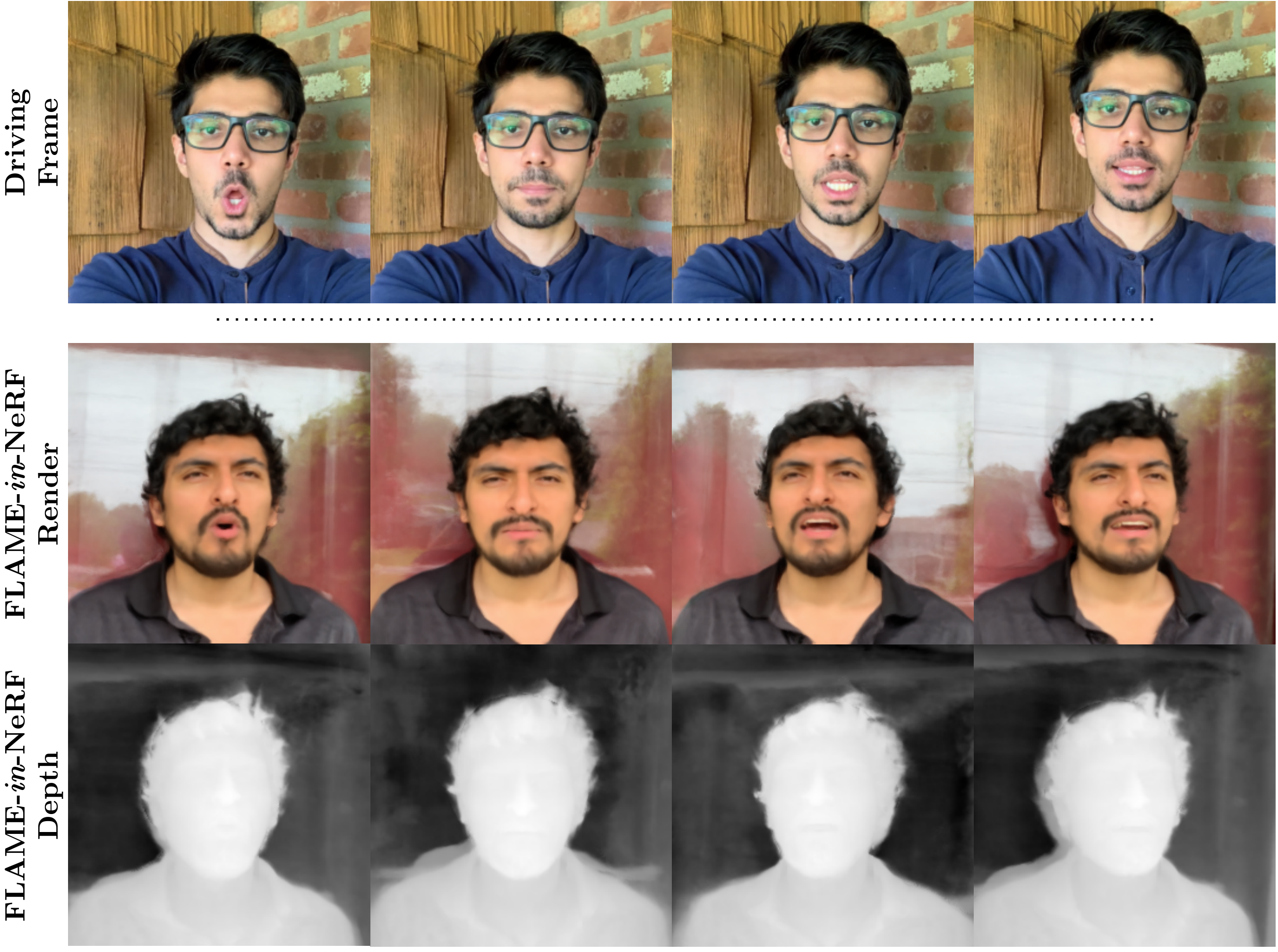}
    
    \caption{{\textbf{Reanimation using \MethodName.} Results of reanimating Subject 4 using a self captured video. In the second row we show the reanimated frames and in the last row we show the rendered depth.   
    }}
    \label{fig:reanim_4}
\end{figure*}

In this section we show more reanimation results of our method using both a self-captured video and a video from the internet. In \fig{reanim_1}, we show the reanimation of four subjects using a self-captured video. As one can see, the reanimated frames have high fidelity to the target expression while simultaneously being individualistic. Similarly, in \fig{reanim_2}, we see the high fidelity of the results to the target expression and the individual characteristics. In \fig{reanim_3} and \fig{reanim_4}, we show results of reanimation and the rendered depth of each reanimated frame.


\subsection{Experimental Configuration}
\label{sect:exp_config}
\begin{figure*}[h]
    \centering
    \includegraphics[width=1.0\linewidth]{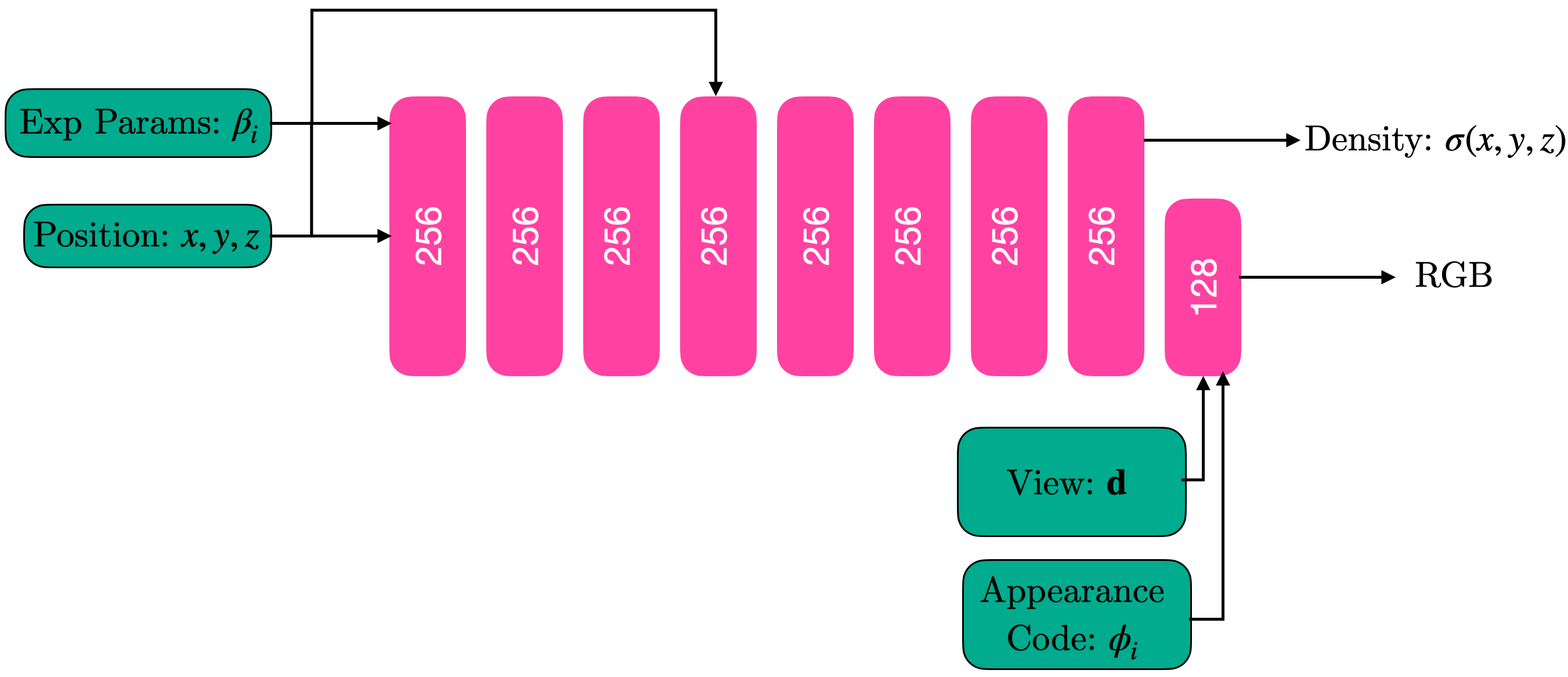}
    
    \caption{{\textbf{Canonical NeRF architecture used in \MethodName.} \MethodName uses the canonical NeRF architecture \cite{nerf} with a hidden layer size of 256. Both the position and view direction are encoded using positional encoding.
    }}
    \label{fig:nerf_net}
\end{figure*}

\begin{figure*}[h]
    \centering
    \includegraphics[width=1.0\linewidth]{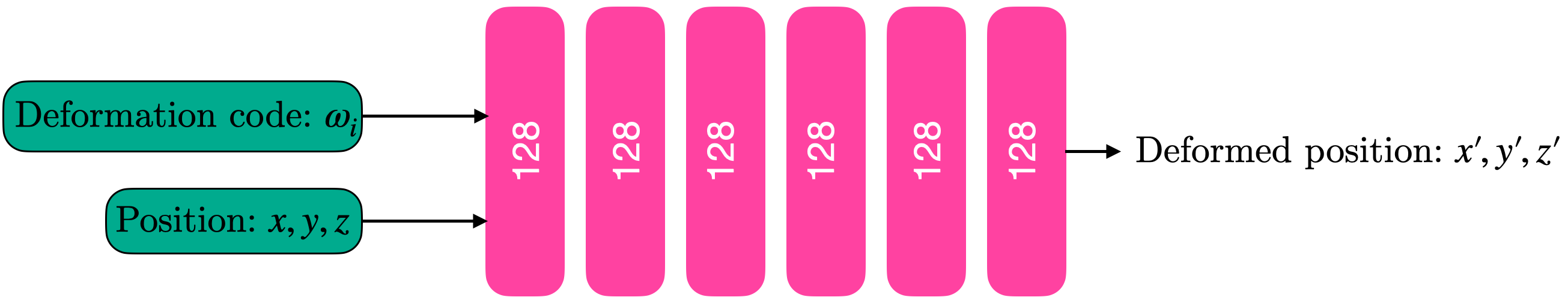}
    
    \caption{{\textbf{Deformation network architecture used in \MethodName.}  \MethodName uses the deformation network architecture from \cite{nerfies} with a hidden layer size of 128. The position is encoded using positional encoding with coarse-to-fine regularization \cite{nerfies} (See Section 3.1 in the paper for details).
    }}
    \label{fig:def_net}
\end{figure*}

All models were trained on 7 Titan RTX GPUs. Both the Coarse and Fine NeRF models \cite{nerf} used 64 points along the ray. The positional is encoded using 10 frequencies while the view is encoded using 4. The Adam optimizer was used for all experiments with a starting learning rate of 1e-3 which was decayed to 5e-4 till the end of training. Coarse-to-fine regularization was applied for 50k epochs (i.e \(N = 50k\), see Eq. 4 of the paper). The network architecture for the canonical NeRF that gives as output the RGB color and density is shown in \fig{nerf_net} and the architecture of the deformation network is shown in \fig{def_net}.

\begin{table}[h]
\begin{tabular}{@{}llllll@{}}
\toprule
Subject                     &Method                     & Epochs Trained       & App Code dim  & Def Code dim  & FRR Coeff\\ \midrule
\multirow{2}{*}{Subject 1} &\MethodName &150000 & 8 & 128 & 1e-1\\
&Nerfies                    & 200000  & 8 & 128 & 1e-2\\ \bottomrule
\multirow{2}{*}{Subject 2} &\MethodName &150000 & 8 & 128 & 10\\
&Nerfies                    & 150000  & 8 & 128 & 10\\ \bottomrule
Subject 3 &\MethodName &80000 & 8 & 128 & 1.0\\\bottomrule
Subject 4 &\MethodName & 80000 & 8 & 128 & 1.0\\\bottomrule
\end{tabular}
\caption{Trainig configuration for all the experiments.}
\label{tab:Subjects_metrics}
\end{table}

\end{document}